\documentclass{article}

\usepackage[preprint]{neurips_2025}

\usepackage[utf8]{inputenc} 
\usepackage[T1]{fontenc}    
\usepackage{newunicodechar}
\newunicodechar{，}{,}
\usepackage{newunicodechar}
\newunicodechar{，}{,}
\usepackage{hyperref}       
\usepackage{url}            
\usepackage{booktabs}       
\usepackage{amsfonts}       
\usepackage{nicefrac}       
\usepackage{microtype}      
\usepackage{xcolor}         
\usepackage{graphicx}
\usepackage{hyperref}       
\usepackage{url}            
\usepackage{booktabs}       
\usepackage{amsfonts}       
\usepackage{nicefrac}       
\usepackage{microtype}      
\usepackage{xcolor}         
\usepackage{multirow}
\usepackage{wrapfig}
\usepackage{pifont}
\usepackage{caption}
\usepackage{wrapfig}
\usepackage{subcaption}
\usepackage{amsmath}   
\usepackage{amsthm}    
\usepackage{xcolor}
\usepackage{tcolorbox}
\usepackage{enumitem}
\tcbuselibrary{skins, breakable}
\usepackage[misc]{ifsym} 
\usepackage{float}     
\newenvironment{teaserfigure}[1][htbp]
{\begin{figure}[#1]\centering}
{\end{figure}}
\newtcolorbox{promptbox}{
    enhanced,
    breakable,
    colback=blue!5!white,
    colframe=blue!75!black,
    arc=3mm,
    boxrule=1.5pt,
    title={\large\bfseries Question Formulation Protocol for Visual Reasoning},
    fonttitle=\sffamily\bfseries,
    attach boxed title to top center={yshift=-2mm},
    boxed title style={
        colback=blue!75!black,
        arc=3mm,
        boxrule=0pt
    }
}

\title{WeThink: Toward General-purpose Vision-Language Reasoning via Reinforcement Learning}

\author{
  \hspace{-0.6cm}{Jie Yang\textsuperscript{\rm 1}~~Feipeng Ma\textsuperscript{\rm 1,3}~~Zitian Wang~~Dacheng Yin\textsuperscript{\rm 1}~~Kang Rong\textsuperscript{\rm 1}~~Fengyun Rao\textsuperscript{\rm 1}}~~Ruimao Zhang\textsuperscript{\rm 2,~{\textrm{\Letter}}} \quad \\ 
  \hspace{-0.8cm}\textsuperscript{\rm 1}WeChat Vision, Tencent~~\textsuperscript{\rm 2}Sun Yat-sen University~~\textsuperscript{\rm 3}University of Science and Technology of China\\ \\
\hspace{-0.7cm} \url{https://github.com/yangjie-cv/WeThink}
}

\begin{document}

\maketitle

\begin{teaserfigure}
    \centering
    \includegraphics[width=1\textwidth]{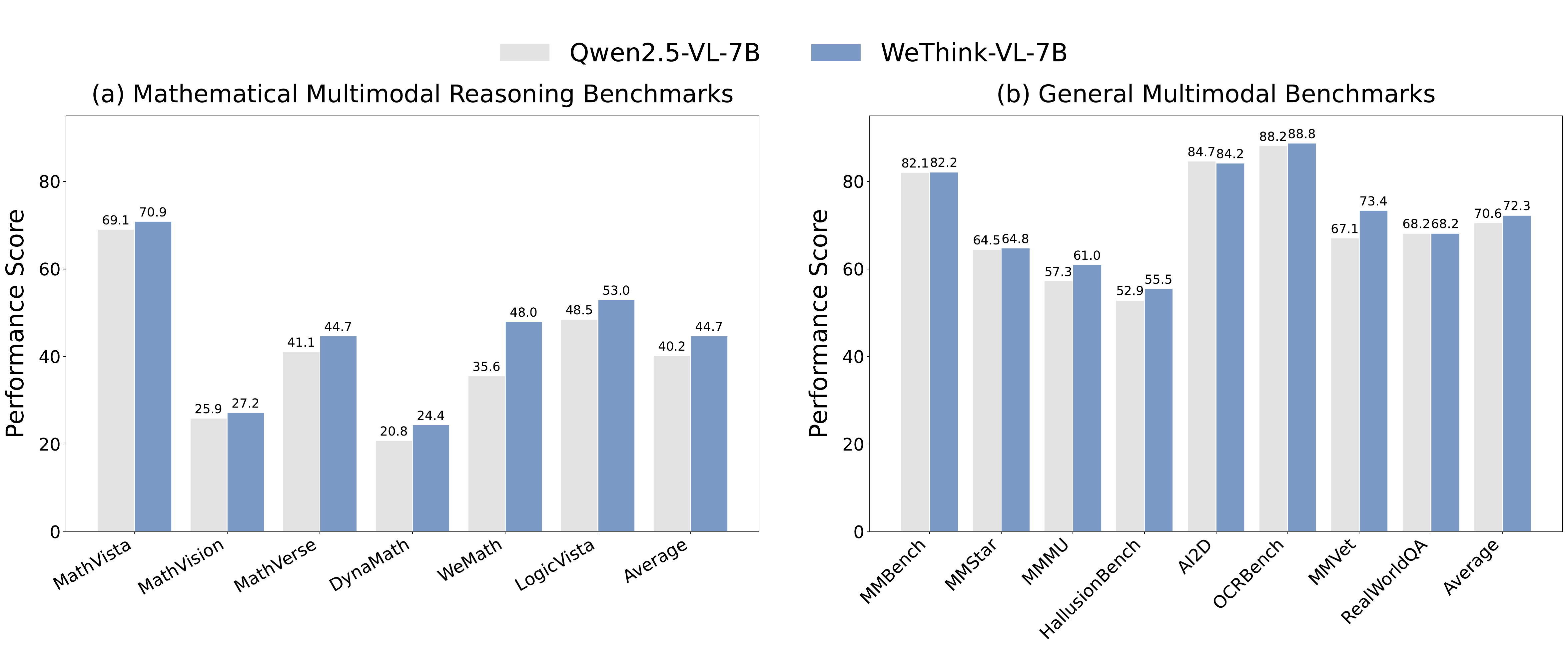}
\caption{WeThink-VL-7B, fine-tuned on Qwen2.5-VL-7B~\cite{bai2025qwen2} through reinforcement learning, shows significant improvements in tasks from mathematical reasoning to general challenges.}
    \label{fig:teaser}
\end{teaserfigure}

\begin{abstract}

Building on the success of text-based reasoning models like DeepSeek-R1, extending these capabilities to multimodal reasoning holds great promise. While recent works have attempted to adapt DeepSeek-R1-style reinforcement learning (RL) training paradigms to multimodal large language models (MLLM), focusing on domain-specific tasks like math and visual perception, a critical question remains: \textit{\textbf{How can we achieve the general-purpose visual-language reasoning through RL?}} To address this challenge, we make three key efforts:  {\textbf{(1)}} A novel \textit{Scalable Multimodal QA Synthesis} pipeline that autonomously generates context-aware, reasoning-centric question-answer (QA) pairs directly from the given images.
{\textbf{(2)}} The open-source \textbf{\texttt{WeThink}} dataset containing over 120K multimodal QA pairs with annotated reasoning paths, curated from 18 diverse dataset sources and covering various question domains.  {\textbf{(3)}} A comprehensive exploration of RL on our dataset, incorporating a hybrid reward mechanism that combines rule-based verification with model-based assessment to optimize RL training efficiency across various task domains.
Across 14 diverse MLLM benchmarks, as shown in Fig.~\ref{fig:teaser}, we demonstrate that our \textbf{\texttt{WeThink}} dataset significantly enhances performance, from mathematical reasoning to diverse general multimodal tasks. Moreover, we show that our automated data pipeline can continuously increase data diversity to further improve model performance.

\end{abstract}

\section{Introduction}
\vspace{-0.4cm}
Visual-Language Reasoning has emerged as a pivotal capability for multimodal large language models (MLLMs), enabling tasks ranging from complex mathematical problem-solving to general visual question answering. Closed-source models like OpenAI's o3~\cite{openai2025o3o4mini} and Kimi k1.5~\cite{team2025kimi} have demonstrated remarkable performance in visual-language reasoning, sparking significant interest within the open-source community. In contrast, recent open-source initiatives such as DeepSeek-R1~\cite{guo2025deepseek} have pioneered text-centric reasoning models by integrating reinforcement learning (RL) with verifiable rewards. However, these models~\cite{guo2025deepseek,chu2025sft,muennighoff2025s1} are inherently constrained to unimodal (\textit{i.e.}, text-only) scenarios, leaving a critical gap in multimodal reasoning capabilities.

Recent works~\cite{chen2025sft,deng2025openvlthinker,zhang2025r1,yang2025r1,wang2025visualprm,huang2025vision,wang2025vl,wei2025skywork,liu2025noisyrollout} have attempted to adapt DeepSeek-R1-style RL training paradigms to MLLMs, focusing primarily on domain-specific tasks like mathematical reasoning and visual perception. Yet, a key question persists: \textbf{\textit{How can we achieve the general-purpose visual-language reasoning through RL?}} Two critical aspects stand out. 

$\bullet$ \textbf{{Diverse Reason-centric Data}}. Recent DeepSeek-R1-style methods rely on pre-collected question-answer (QA) datasets for cold-start supervised fine-tuning (SFT) with Chain-of-Thought (CoT) annotations or for reformulating answers to calculate accuracy rewards in RL. However, these QA pairs often lack the multi-step reasoning needed for robust visual-language reasoning. Additionally, some methods are dependent on domain-specific question types, which limits their scalability across various domains. To enable more general visual-language reasoning, it's important to use diverse, reason-focused data from a wide range of domains and contexts.
    
$\bullet$ \textbf{{RL with Hybrid Rewards}}. While rule-based rewards (e.g., answer verification for mathematical problems) are effective in specific domains, they struggle to capture the complexity of general multimodal scenarios, where answers can be subjective or context-dependent. This underscores the need for a hybrid reward system that combines both rule-based and model-based strategies, offering more nuanced, context-sensitive feedback to enable RL-trained MLLMs to handle diverse task domains.

To address the data aspect, we propose an novel \textbf{\textit{Scalable Multimodal QA Synthesis}} pipeline that can autonomously generate context-aware, reason-centric questions paired with verifiable answers directly from the given images. It can benefit from diverse data sources, including open-source datasets and various resources across the Internet, enabling the continuous enhancement of data diversity.
To further contribute to the field, we open-source the \textbf{\texttt{WeThink}} dataset, which contains over 120K multimodal QA pairs with explicit reasoning paths. Curated from 18 distinct public image datasets, \textbf{\texttt{WeThink}} encompasses a broad range of question domains and types, requiring integrated abilities such as \textit{reasoning}, \textit{OCR}, \textit{recognition}, \textit{math}, \textit{knowledge}, and \textit{spatial awareness}, thereby enhancing general multimodal reasoning capabilities.

Building upon our dataset, we conduct a comprehensive exploration of RL on \textbf{\texttt{WeThink}}, introducing a hybrid reward mechanism that integrates rule-based verification with model-based evaluation to enhance RL training efficiency across diverse task domains. Through extensive experiments, we present four key findings:
\textbf{(1)} SFT with CoT supervision on our dataset enhances the performance of less optimized model (e.g., Qwen2-VL-7B), yielding an average improvement of 3.5\% across six mathematical reasoning benchmarks.
\textbf{(2)} In our scenarios, direct RL fine-tuning on Qwen2.5-VL-7B~\cite{bai2025qwen2} is sufficient and even outperforms cold-start supervised fine-tuning followed by RL.
\textbf{(3)} With our dataset, increasing the diversity of question domains through RL fine-tuning leads to significant improvements across tasks, from mathematical reasoning to general tasks, as shown in Fig.~\ref{fig:teaser}.
\textbf{(4)} The scalability of our data pipeline enables continuous collection of diverse images from the Internet, further enhancing model performance.

In summary, the contributions of this work are three-fold:

 $\diamond$ \textbf{Automated Data Generation Pipeline}: We propose an novel \textit{Scalable Multimodal QA Synthesis} pipeline that autonomously generates context-aware, reason-centric questions paired with verifiable answers directly from the given images.
 
 $\diamond$ \textbf{Diverse Reason-centric Dataset}: We open-source the \textbf{\texttt{WeThink}} dataset, containing over 120K multimodal QA pairs with explicit reasoning paths, curated from 18 distinct public image datasets. It spans various question domains and types, enhancing multimodal reasoning capabilities in models.

 $\diamond$ \textbf{General-Purpose Visual-Language Reasoning Models}: We conduct a comprehensive exploration of reinforcement learning on our dataset and present a series of models that improve performance in tasks from mathematical reasoning to general challenges. Additionally, we show that our pipeline's scalability, driven by increasing data diversity, further improves model performance.

\section{Related Work}

\subsection{Multimodal Large Language Models (MLLMs)}
Recent years have witnessed significant advancements in Multimodal Large Language Models (MLLMs), which augment traditional Large Language Models (LLMs) by enabling them to process and comprehend information from diverse modalities, including text, images, audio, and video~\cite{liu2023visual,barrault2023seamlessm4t,chen2022wavlm,li2023videochat,zhang2023video}.
The rapid evolution in this field is evidenced by the development of numerous open-source models, such as MiniGPT-4~\cite{zhu2023minigpt}, MiniCPM-V~\cite{yao2024minicpm}, CogVLM~\cite{wang2024cogvlm}, ShareGPT4V~\cite{chen2024sharegpt4v}, Qwen-VL~\cite{Qwen-VL,wang2024qwen2,bai2025qwen2}, LLaVA~\cite{liu2023visual,liu2024improved,li2024llava}, and InternVL~\cite{chen2024internvl,chen2024far,chen2024expanding}, alongside prominent closed-source models like Gemini~\cite{team2023gemini,team2024gemini}, GPT-4o~\cite{hurst2024gpt}, Claude~\cite{anthropic_claude_apa}, and Grok~\cite{xai_grok}. These efforts highlight ongoing progress in architectural designs, pre-training strategies, and instruction tuning techniques.
Despite these strides, fostering robust reasoning capabilities across diverse domains and tasks continues to be a primary area of research.

\subsection{Chain-of-Thought Prompting for Multimodal Reasoning}
Chain-of-Thought (CoT) prompting, a technique that significantly enhances the reasoning capabilities of Large Language Models (LLMs) by guiding them to articulate intermediate inferential steps prior to delivering a final answer~\cite{wei2022chain, kojima2022large}, has been naturally and effectively extended to the multimodal domain. 
In Multimodal Large Language Models (MLLMs), the application of CoT not only demonstrably improves performance on complex reasoning tasks but also offers enhanced interpretability into the model's intricate decision-making processes~\cite{zhang2023multimodal,lu2023chameleon,luo2024layoutllm}.
A variety of strategies have been developed to elicit, generate, and leverage CoT reasoning in MLLMs. These include designing structured reasoning templates or programmatic approaches to systematically guide the CoT process~\cite{yang2023mm,zhang2024cocot,mitra2024compositional,zheng2023ddcot,ni2024visual,chen2024visual} and Supervised Fine-Tuning (SFT) using datasets enriched with multimodal CoT examples~\cite{xu2024llava,dong2024insight,thawakar2025llamav}.
As these strategies often generate pre-defined or limited thought processes, there is a growing focus on integrating them with reinforcement learning, to encourage exploration of diverse problem-solving strategies, and ultimately develop more sophisticated and genuinely intelligent multimodal reasoning capabilities.

\subsection{Reinforcement Learning for Multimodal Reasoning}

Reinforcement Learning (RL) has emerged as a transformative approach for enhancing reasoning capabilities in Multimodal Large Language Models (MLLMs)~\cite{zhou2025reinforced}. The integration of RL, particularly Reinforcement Learning from Human Feedback (RLHF)~\cite{ouyang2022training} or rule-based reward mechanisms (R1-style)~\cite{guo2025deepseek}, aims to align MLLM outputs with desired reasoning patterns and task objectives.
Recently, several works have successfully adapted and extended the R1-style RL training paradigm into MLLMs. These efforts have primarily focused on exploring how R1-style RL can enhance MLLM capabilities in math-centric multi-modal reasoning~\cite{chen2025sft,deng2025openvlthinker,zhang2025r1,yang2025r1,wang2025visualprm,huang2025vision,wang2025vl,wei2025skywork,liu2025noisyrollout} and various specific downstream tasks. For instance, researchers have applied these RL techniques to improve scene graph understanding~\cite{li2025relation,chen2025compile}, visual-spatial reasoning~\cite{zhao2025embodied,liao2025improved,zhang2025embodied,zhou2025r1}, referring expression comprehension~\cite{yu2025perception,shen2025vlm,liu2025visual,deng2025boosting,liu2025seg}, and visual counting~\cite{wang2025crowdvlm,tan2025reason,liu2025othink}. While these methods have demonstrated promising results within their respective scopes, few have focused on leveraging RL to broadly enhance the general multi-modal understanding and reasoning abilities of MLLMs.
 \section{WeThink Dataset with Scalable Multimodal QA Synthesis}

\begin{figure}[t]
    \centering
\includegraphics[width=\textwidth]{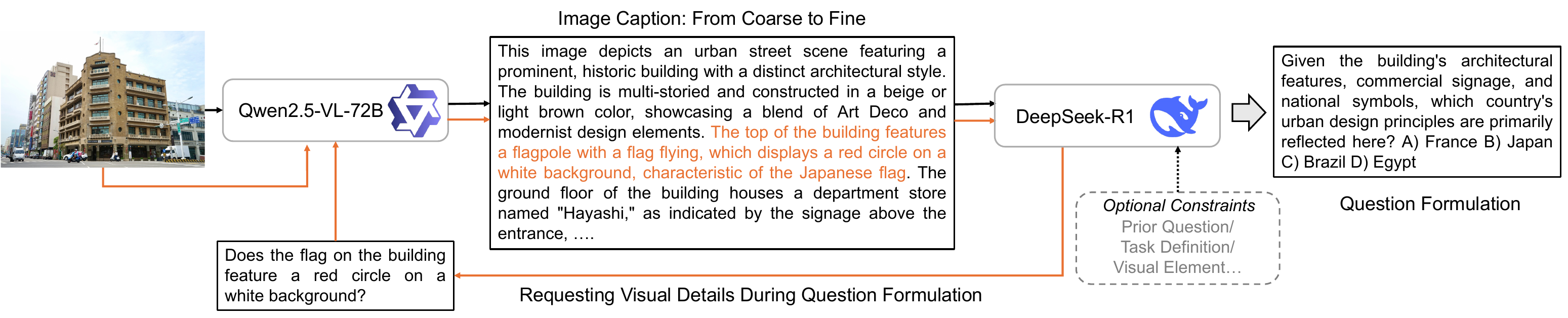} 
\caption{The automatic process of question formulation for a given image. As illustrated by the \textcolor{orange}{orange} line, based on the coarse description provided by Qwen2.5-VL-72B, DeepSeek-R1 needs to request additional visual details (\textcolor{orange}{orange} text) through multi-turn conversations with Qwen2.5-VL-72B, thus facilitating the generation of context-aware, reasoning-centric questions. We also highlight that the process can condition various constraints through prompts, such as prior questions (if available), task definition, and visual focus, to control the type and focus of the questions.}
    \label{fig:question_formulation_process}
        \vspace{-0.5cm}
\end{figure}

This section presents an automated \textit{Scalable Multimodal QA Synthesis} pipeline, designed to generate context-aware, reasoning-centric question-answer (QA) pairs from the given images. We also introduce \textbf{\texttt{WeThink}}, a dataset carefully curated to encompass diverse question domains, types, and integrated abilities. Below, we describe the processes of data collection, question formulation, answer construction, and quality control, and conclude by presenting the data characteristics of \textbf{\texttt{WeThink}}.

\subsection{Data Collection}
Our pipeline is designed to autonomously generate high-quality QA pairs directly from the given images. These images can come from open-source datasets or various sources across the Internet.
Here, to show its effectiveness, we collect open-source images to publicly release the generated QA pairs. Specifically, we sample images from 18 distinct datasets, as used in LLaVA-CoT~\cite{xu2024llava}, shown in Tab.~\ref{tab:dataset-en}. These datasets cover diverse image categories, ensuring variety and complexity in the generated QA pairs, including general images (COCO~\cite{lin2014microsoft}, SAM-1B~\cite{kirillov2023segment}, Visual Genome~\cite{krishna2017visual}, GQA~\cite{hudson2019gqa}, PISC~\cite{li2017dual}, LLaVA~\cite{liu2023visual}), text-intensive images (TextVQA~\cite{singh2019towards}, ShareTextVQA~\cite{chen2024sharegpt4v}, DocVQA~\cite{mathew2021docvqa}, OCR-VQA~\cite{mishra2019ocr}, ChartQA~\cite{masry2022chartqa}), scientific and technical images (GeoQA+~\cite{cao2022augmented}, ScienceQA~\cite{lu2022learn}, AI2D~\cite{kembhavi2016diagram}, CLEVR-Math~\cite{lindstrom2022clevr}), and images related to art and culture contexts (WikiArt~\cite{saleh2015large,chen2024sharegpt4v}, Web-Landmark~\cite{kreutzer2022quality,chen2024sharegpt4v} Web-Celebrity~\cite{kreutzer2022quality,chen2024sharegpt4v}). 

\subsection{Question Formulation}
\label{sec:question_formulation}
Based on the collected images, we aim to generate context-aware, reasoning-centric questions. The straightforward pipeline involves collaboration between two powerful models, Qwen2.5-VL-72B~\cite{bai2025qwen2} and DeepSeek-R1~\cite{guo2025deepseek}, to analyze images and generate questions. In this workflow, the visual-language model Qwen2.5-VL-72B first provides a detailed description of the image, after which language-only model DeepSeek-R1 analyzes the description, reflects on its content, and synthesizes relevant questions based on the analysis.
However, two critical challenges arise: \textbf{(1)} incomplete visual understanding by Qwen2.5-VL-72B, \textbf{(2)} uncontrolled complexity and reasoning focus of question generation by DeepSeek-R1. To address these issues, we carefully design the question formulation process with two core strategies: \textit{Multi-turn Information Refinements} and \textit{Ability Synergy Constraints}, along with \textit{Optional Contextual Constraints}.

\textbf{Multi-turn Information Refinements.} As shown in Fig.~\ref{fig:question_formulation_process}, given that image descriptions provided by Qwen2.5-VL-72B may sometimes be insufficient or erroneous, we implement a multi-turn information refinement mechanism to address such shortcomings, including three stages:
\vspace{-0.2cm}
\begin{itemize}
\item  \textbf{Coarse Description Generation}: Qwen2.5-VL-72B extracts global features from the input image and generates an initial description that provides a broad overview of the main visual elements. This serves as the semantic anchor for subsequent multi-turn dialogues.

\item \textbf{Dynamic Detail Mining}: To generate context-aware, reasoning-centric questions, DeepSeek-R1 identifies information gaps based on the initial coarse description. It then generates follow-up questions to request more detailed visual information from Qwen2.5-VL-72B. This process ensures that the questions address all relevant aspects of the image, including the reasoning needed for the final question formulation.
\item \textbf{Context Integration}: As each piece of supplementary information is gathered during the multi-turn dialogue, it is integrated into the evolving description. Qwen2.5-VL-72B records this information and synthesizes it into a final, fine-grained description, which serves as the basis for generating a comprehensive and contextually aware question.
\end{itemize}
\vspace{-0.2cm}
\textbf{Ability Synergy Constraints.} Inspired by MM-Vet benchmark~\cite{yu2023mm} that evaluates the model's integrated capabilities, we propose to incorporate multi-ability constraints into the question formulation phase. This approach aims to create more complex questions that better reflect the model's ability to apply and combine the skills learned during training.  Specifically, we present formulation protocol in Prompt.~\ref{sec:output_specifications}-III, with more detail illustrations in Appendix~\ref{sec:data_supment}.
The formulation enforces mandatory \textcolor{blue}{reasoning} capability combined with at least one complementary ability from other five-dimensional taxonomy: \textit{1) Recognition}: General visual recognition (e.g., objects, attributes, scenes, counting, or high-level computer vision tasks); \textit{2) Knowledge}: Use of social/visual commonsense, encyclopedic knowledge, or recent/contextual information; \textit{3) OCR}: Reading and reasoning over visible text (e.g., scene text, handwritten text, or embedded text in objects); \textit{4) Spatial Awareness}: Understanding spatial relationships (e.g., object positions, directional/distance logic, layout analysis); \textit{5) Math}: Performing arithmetic operations, solving equations, or interpreting math-specific notation. Overall, the above mechanism ensures questions inherently require:
\vspace{-0.2cm}
\begin{itemize}
\item  \textbf{Cross-modal Reasoning Chains}: Minimum two explicit reasoning chains with comprehensive image analysis
\item  \textbf{Semantics-Driven Ability Selection}: Automated activation of relevant abilities based on image content information
\vspace{-0.2cm}
\end{itemize}
\begin{figure}[t]
    \centering
    \includegraphics[width=\textwidth]{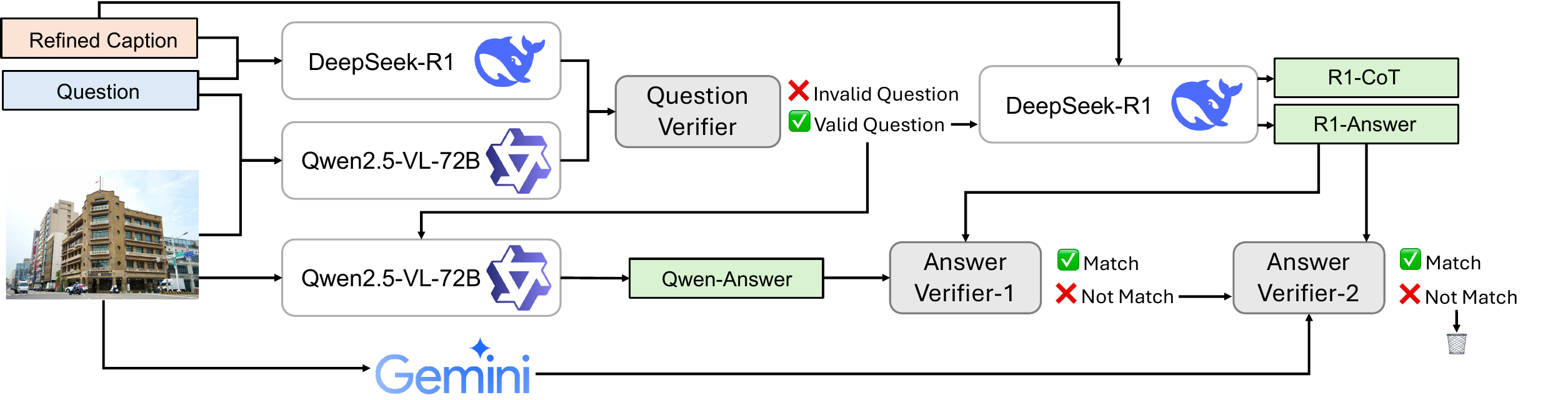} 
\caption{The automatic process of answer construction and quality control. First, DeepSeek-R1 filters out forced and open-ended questions to ensure they are verifiable. Then, using the refined caption and the valid question, DeepSeek-R1 generates a chain of thought and an answer. At the same time, Qwen2.5-VL-72B generates an answer based on the image. If their answers match, the result is kept; if not, Gemini re-evaluates DeepSeek-R1's answer, discarding incorrect responses and keeping only the correct one with its chain of thought.}
    \label{fig:answaer_formulation_process}
    \vspace{-0.5cm}
\end{figure}

For instance, a question requiring \textcolor{green!50!black}{object recognition} + \textcolor{orange}{historical knowledge} might ask: ``Given the architectural style of the building's columns shown, what historical period does this structure represent?'' This combines visual feature extraction with architectural history knowledge.

\textbf{Optional Contextual Constraints.} To achieve better precision and control over the generated questions, particularly in terms of their type and focus, we can optionally condition the generation process with contextual constraints. Fortunately, open-source collections often include QA pairs, and web images typically come with textual descriptions or captions. For instance, these may include prior questions to guide related queries, a clear task definition to direct the goal, and visual cues to highlight specific areas of the image. By leveraging these optional constraints, the pipeline can generate more targeted, relevant, and user-intended questions in a controlled manner.

\subsection{Answer Construction and Quality Control}
For the generated questions, the process of answers construction follows a structured approach consisting of three key stages: {\textit{Preliminary Question Filtering}}, {\textit{Answer Construction and Quality Control}}, as well as {\textit{CoT  Refinement}}. As shown in Fig.~\ref{fig:answaer_formulation_process}, each stage is carefully designed to ensure the accuracy and reliability of the answers through a multi-model verification framework.

\textbf{Preliminary Question Filtering.}
Considering the instability of question formulation, our first step is to filter out questions that are unverifiable, ambiguous, or irrelevant to the image. In practice, we perform two rounds of verification. Firstly, DeepSeek-R1 uses a refined image caption to filter out invalid questions. Then, we also apply the visual-language model Qwen2.5-VL-72B to analyze the image further and filter out additional invalid questions.
The remaining questions are categorized into three types: multiple-choice (MC), fill-in-the-blank (FIB), and descriptive (DES).

\textbf{Answer Construction and Quality Control.}  The next stage is to generate and verify answers across different question types. For \textbf{\textit{MC}} and \textbf{\textit{FIB}} questions, which can be verified using rules, DeepSeek-R1 generates answers based on a refined image description, while Qwen2.5-VL-72B generates answers from the image content. These answers are then compared for alignment, and if they match, they are considered reliable. In cases of discrepancies, a secondary evaluation by another powerful visual language model Gemini 2.5 Pro~\cite{team2024gemini} is performed to re-evaluate and discard incorrect answers. For \textbf{\textit{DES}} questions, which often require longer, more detailed answers to stimulate reasoning and interpretation of complex visual data, DeepSeek-R1 generates the final answer, and Qwen2.5-VL-72B directly verifies its correctness. If the answer is confirmed, it is retained; otherwise, Gemini re-assesses and filters out incorrect QA pairs.

\textbf{CoT Refinement.} 
During the answer construction process, DeepSeek-R1 naturally generates a chain-of-thought (CoT) for each question. However, we observed that these CoTs are often overly lengthy and contain redundancies. Our subsequent experiments also revealed that these CoTs are suboptimal for both direct SFT and as cold start data for RL training. To address this, we refine the CoTs by incorporating both the image and the final answer into the QwenVL2.5-72B. This refinement process yields more concise CoTs, allowing us to more effectively investigate how CoT quality influences both SFT and its role as cold-start data for RL training.

\begin{figure}[t]
\centering
\begin{minipage}{0.35\textwidth} 
\centering
\captionof{table}{The distribution analysis of image types from \texttt{WeThink}.}
\resizebox{\textwidth}{!}{%
\begin{tabular}{lcc}
\toprule
\textbf{Image Type} & \textbf{Source Dataset} & \textbf{Images} \\
\midrule
\multirow{6}{*}{General Images} 
 & COCO & 30786 \\
  & SAM-1B & 12014 \\
   & Visual Genome & 4414 \\
 & GQA & 3483 \\
 & PISC & 1148 \\
  & LLaVA & 150 \\
\midrule
\multirow{4}{*}{Text-Intensive Images}
 & TextVQA & 17571 \\
 & ShareTextVQA & 429 \\
 & DocVQA & 5805 \\
 & OCR-VQA & 6485 \\
  & ChartQA & 22865 \\
\midrule
\multirow{5}{*}{Scientific \& Technical}
 & GeoQA+ & 4607 \\
 & ScienceQA  & 3236 \\
  & AI2D & 12024 \\
    & CLEVR-Math  & 434 \\
\midrule
\multirow{3}{*}{Art \& Culture}
 & WikiArt & 401 \\
 & Web-Landmark & 256 \\
 & Web-Celebrity & 319 \\
\bottomrule
\end{tabular}%
}
\label{tab:dataset-en}
\end{minipage}%
\hfill 
\begin{minipage}{0.6\textwidth} 
\centering
\includegraphics[width=\textwidth]{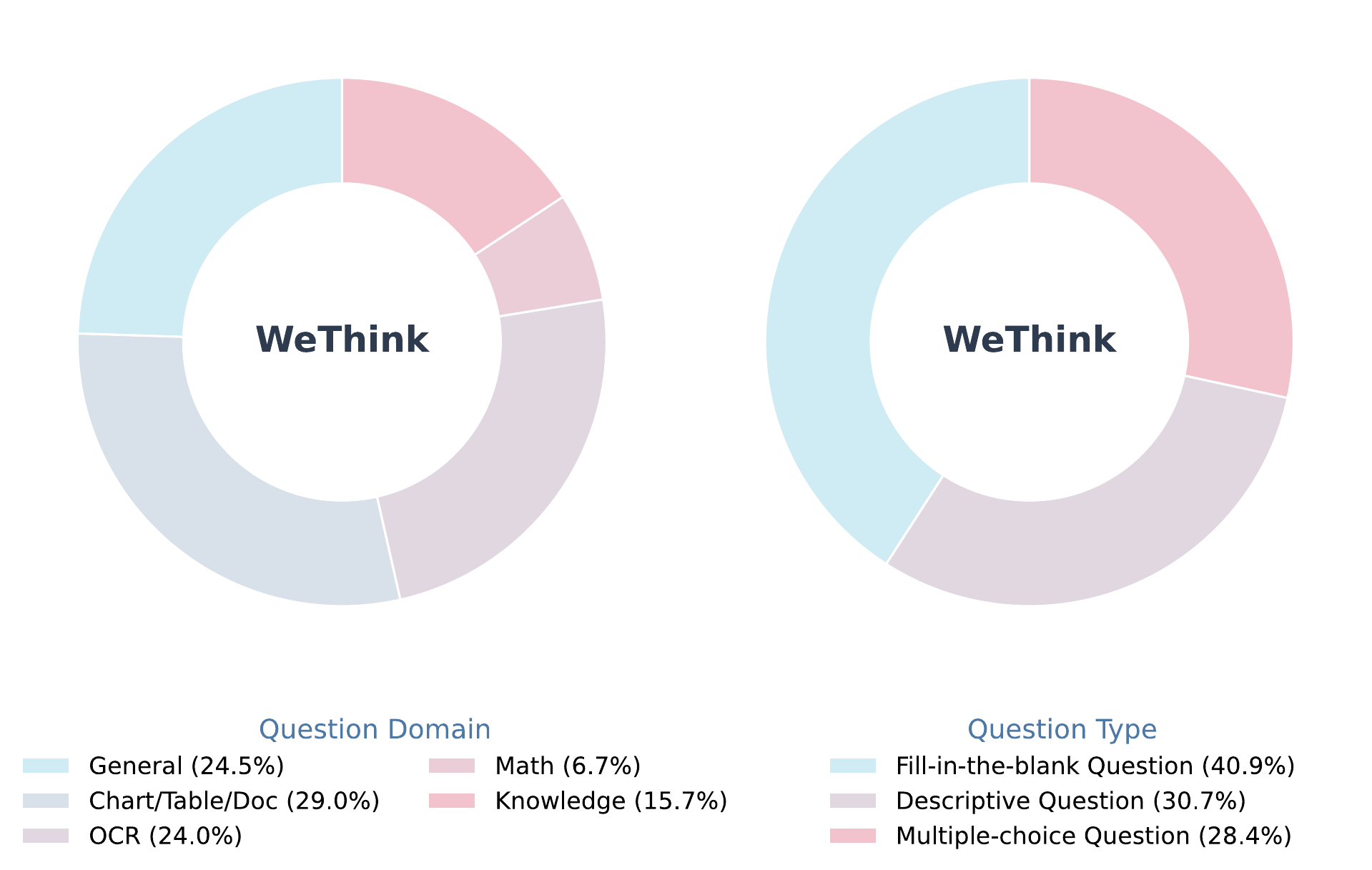}
\caption{The distribution analysis of question domains and types from \texttt{WeThink}.}
\label{fig:question-dist}
\end{minipage}
    \vspace{-0.5cm}
\end{figure}

\begin{figure}[t]
    \centering
    \begin{subfigure}[b]{0.5\textwidth}
    \vspace{0.2cm}
        \includegraphics[width=\linewidth]{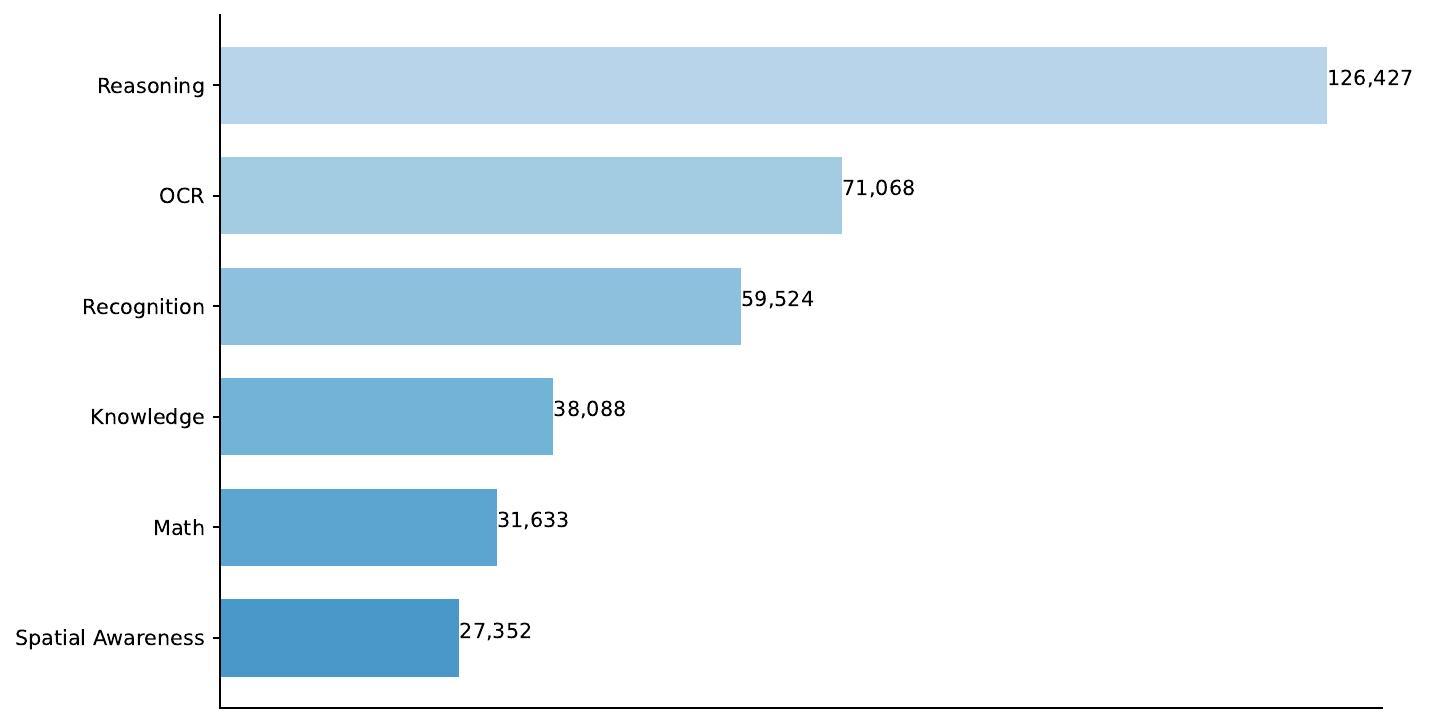}
           \caption{The distribution of each ability}
        \label{subfig:dist}
    \end{subfigure}
    \hfill
    \begin{subfigure}[b]{0.48\textwidth}
\includegraphics[width=\linewidth]{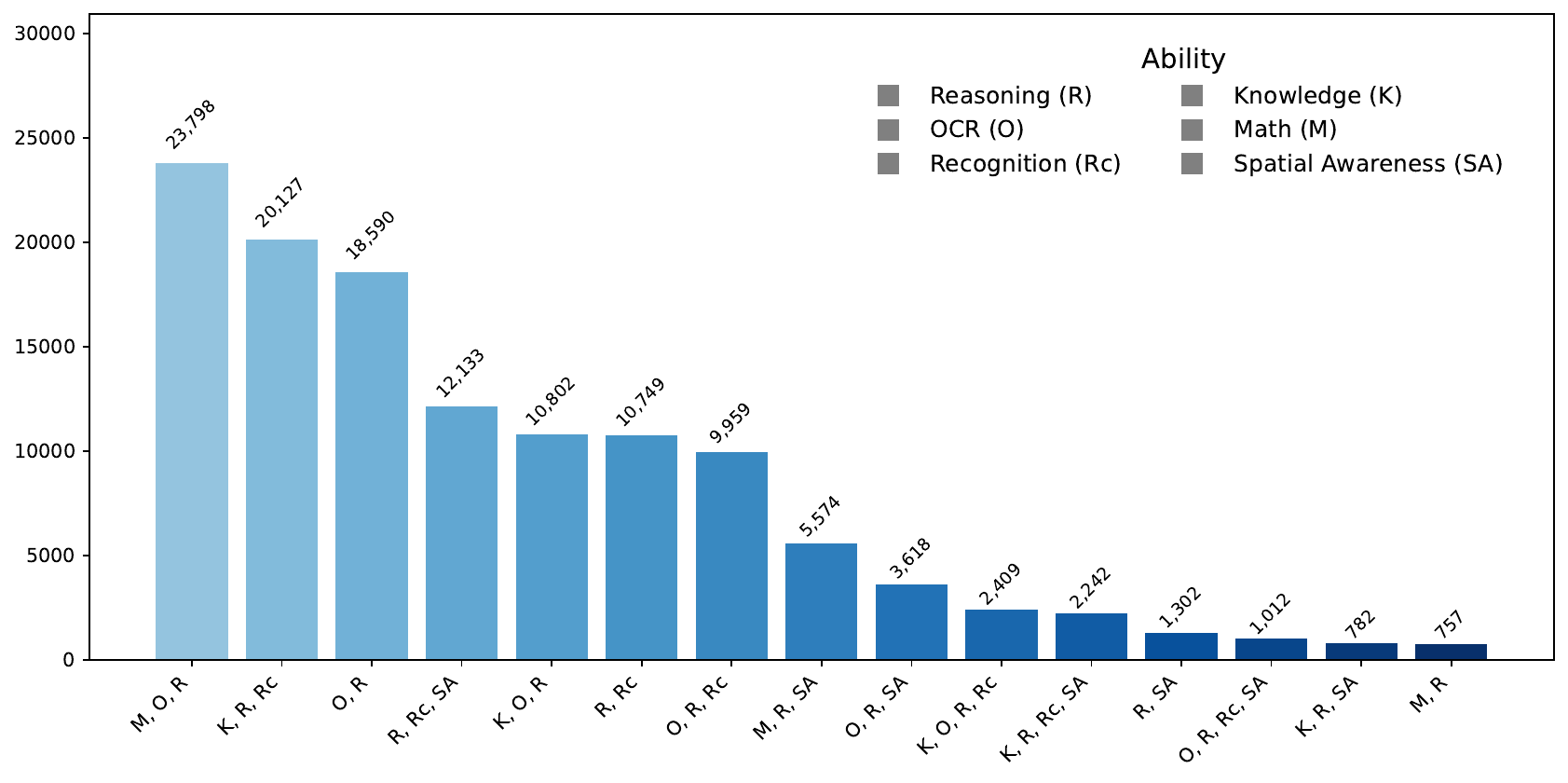}
        \caption{The distribution of top-15 ability combinations}
        \label{subfig:combo}
    \end{subfigure}
    \caption{We show the difficutity distribution of question sample in the  WeThink dataset.}
    \label{fig:ability}
    \vspace{-0.6cm}
\end{figure}

\subsection{Data Characteristics}
Based on the above processes, we have constructed a new dataset named \texttt{WeThink} from open-source images, which offers over \textbf{120K} comprehensive multimodal question-answer pairs with explicit reasoning paths. As a diverse and scalable resource, \texttt{WeThink} was carefully curated to encompass a broad range of question domains, types, and required integrated abilities. To better understand the dataset's structure and focus, we will analyze the following two critical aspects:

\textbf{Question Distribution.} We use Qwen2.5--VL-72B to categorize each question into five groups: general, math, chart/table/doc, knowledge, and OCR. As shown in Fig.~\ref{fig:question-dist}, these categories are fairly balanced, with math being the least common. The questions are also divided into three types: multiple-choice, fill-in-the-blank, and descriptive. The first two types are suitable for RL training with rule-based rewards, while the third is used for RL training with model-based rewards. Each question type is designed for different scenarios and includes reasoning paths.
 
\textbf{Required Ability Distribution.}
The questions in the \texttt{WeThink} dataset are designed to integrate multiple abilities, thereby controlling the difficulty level and stimulating training across various model capabilities. As shown in the Fig.~\ref{subfig:dist}, the core ability is reasoning, as every sample activates this ability. However, other abilities are also triggered depending on the semantic contents of the images. Additionally, each sample engages at least two abilities simultaneously, showcasing top-15 ability combinations, as illustrated in Fig.~\ref{subfig:combo}. Notably, these combinations follow a long-tail distribution, with some ability combinations being rarer than others.

\section{Experiments}

\begin{table}[t]
\centering
\caption{The impact of fully supervised fine-tuning on \texttt{WeThink}. * denotes the model results reproduced by us. Numbers with underlines indicate models compared to. We highlight the best average results in \textbf{bold}, improvements in \textcolor{red}{red}, and decreases in \textcolor{blue}{blue}.}
\resizebox{\textwidth}{!}{ 
\begin{tabular}{l | c c c c c  c |c}
\toprule[1.5pt]
Model & MathVista & MathVision & MathVerse &  DynaMath & WeMath & LogicVista & Average \\ 
\midrule
Qwen2-VL-7B &61.6 &	19.2&	25.4	&11.0	&22.3&	33.3 & 28.8\\
Qwen2-VL-7B* &61.8 &	19.0&	25.6	&11.0	&21.4 &	34.7 & \underline{\textbf{28.9}}\\ 
Qwen2-VL-7B-SFT (R1-CoT) & 56.6 (\textcolor{blue}{-5.2}) & 17.0 (\textcolor{blue}{-2.0}) & 25.0 (\textcolor{blue}{-0.6}) & 16.0 (\textcolor{red}{+5.0}) & 18.5 (\textcolor{blue}{-2.9})& 35.8 (\textcolor{red}{+1.1}) & 28.2 (\textcolor{blue}{-0.6})  \\
Qwen2-VL-7B-SFT (Refined-CoT) & 59.5 (\textcolor{blue}{-2.3}) & 20.3 (\textcolor{red}{+1.3}) & 32.9 (\textcolor{red}{+7.3}) & 15.2 (\textcolor{red}{+5.2}) & 25.0 (\textcolor{red}{+3.6}) & 41.2 (\textcolor{red}{+6.5}) & 32.4 (\textcolor{red}{+3.5})\\
        \midrule \midrule
Qwen2.5-VL-7B & 68.1	&25.4	&41.1	&21.8	&36.2	&47.9&  40.1 \\
Qwen2.5-VL-7B* & 69.1 & 25.9 & 41.1 & 20.8 & 35.6 & 48.5 & \underline{\textbf{40.2}}\\
Qwen2.5-VL-7B-SFT (R1-CoT) & 61.4 (\textcolor{blue}{-7.7}) & 19.7 (\textcolor{blue}{-6.2}) & 31.5 (\textcolor{blue}{-9.6}) & 21.6 (\textcolor{blue}{-0.8}) & 21.7 (\textcolor{blue}{-13.9}) & 34.2 (\textcolor{blue}{-14.3}) & 31.7 (\textcolor{blue}{-8.5}) \\
Qwen2.5-VL-7B-SFT (Refined-CoT) & 63.9 (\textcolor{blue}{-5.2}) & 24.0 (\textcolor{blue}{-1.9}) & 40.6 (\textcolor{blue}{-0.5}) & 19.0 (\textcolor{blue}{-1.8}) & 32.4 (\textcolor{blue}{-3.2})& 40.0 (\textcolor{blue}{-8.5}) & 36.7 (\textcolor{blue}{-3.5})\\
\bottomrule[1.5pt]
\end{tabular}
}
\vspace{-0.4cm}
\label{tab:sft}
\end{table}

\begin{figure}[t]  
    \centering
    \includegraphics[width=\textwidth]{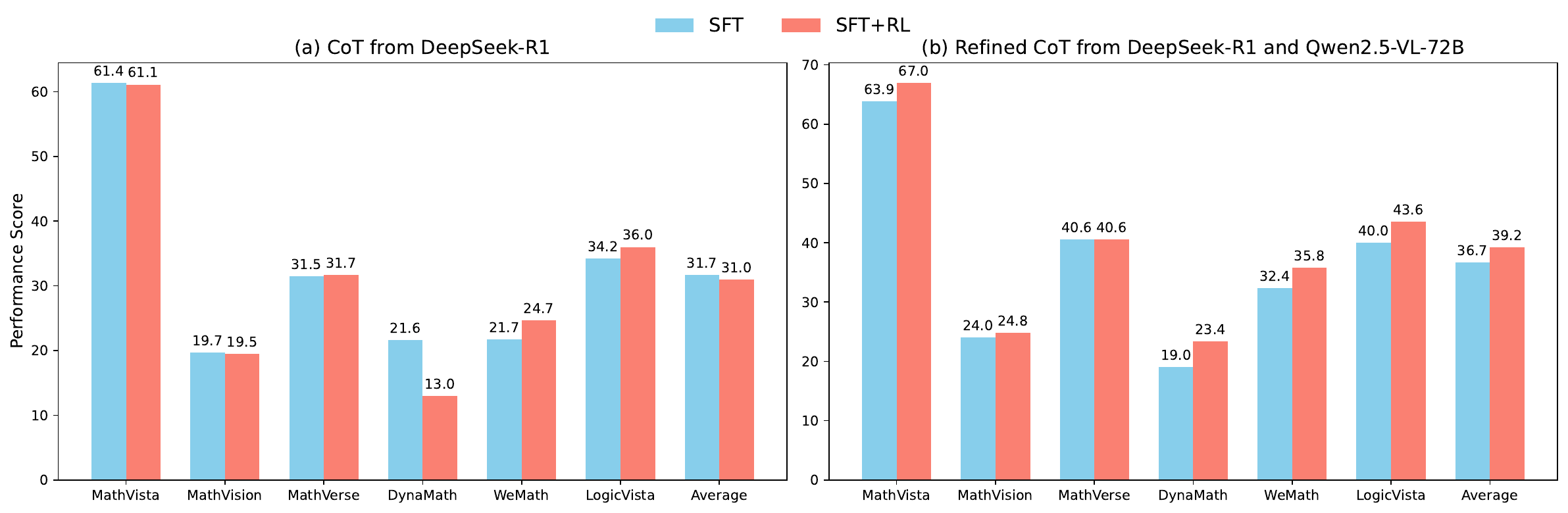}  
    \vspace{-0.5cm}
    \caption{The impact of cold start data quality on performance.}
    \label{fig:sft_vs_rl_comparison} 
    \vspace{-0.5cm}
\end{figure}

\subsection{Benchmarks \& Evaluation}
\label{sec:benchmarks}
To comprehensively evaluate general multi-modal understanding and reasoning capabilities of our models, we conduct experiments across diverse 14 benchmarks, including \textbf{(1) Mathematical Reasoning:} We evaluate on six benchmarks: MathVista~\cite{lu2023mathvista}, MathVision~\cite{wang2024measuring}, MathVerse~\cite{zhang2024mathverse}, DynaMath~\cite{zou2024dynamath}, WeMath~\cite{qiao2024we}, and LogicVista~\cite{xiao2024logicvista},
\textbf{(2) General Multimodal Understanding:} This includes MMMU~\cite{yue2024mmmu} (college-level tasks), MMBench v1.1~\cite{liu2024mmbench} (bilingual multimodal benchmark), MMVet~\cite{yu2023mm} (integrated tasks), MMStar~\cite{chen2024we} (multimodal understanding), HallusionBench~\cite{guan2024hallusionbench} (multimodal hallucination), AI2D~\cite{kembhavi2016diagram} (science diagrams), and OCRBench~\cite{liu2024ocrbench} (OCR capabilities),  RealWorldQA~\cite{XAI2024} (real-world comprehension).
The above benchmarks are available on the OpenCompass MLLM Leaderboard. 
To maintain fairness and reproducibility, we evaluate our models using VLMEvalKit~\cite{duan2024vlmevalkit}, which is an open-source toolkit for MLLM evaluation. \textbf{\textit{We present the details of the benchmarks and implementation of the evaluation in Sec.~\ref{sec:eval}.}}

\subsection{Exploring Visual-Language Reasoning with Supervised Fine-Tuning on \texttt{WeThink}}
\label{sec:sft_exp}
\textbf{Methods \& Implementation Details.} 
Inspired by recent works~\cite{xu2024llava} that explore explicit reasoning processes through chain-of-thought (CoT) prompting during supervised fine-tuning (SFT), we conduct experiments using over 120K CoT-annotated diverse QA pairs from our \texttt{WeThink} dataset for direct SFT. \textit{\textbf{We present the problem definition and detailed optimization formula of SFT in Appendix~\ref{sec:problem} and \ref{sec:SFT}.}} We investigate the impact of two types of CoT (\textit{i.e.}, the original R1-CoT and the refined CoT) on powerful open-source models. In practice, we perform full-parameter fine-tuning for 1 epoch using 8 NVIDIA H20 GPUs on two instruction-tuned models: Qwen2-VL-7B and Qwen2.5-VL-7B. The system prompt is shown in Appendix Tab.~\ref{system}.

\textbf{Results.} 
As shown in Table~\ref{tab:sft}, the experimental results reveal two key findings:
(1) \textbf{\textit{The quality of the SFT CoT is critical.}} The original R1-CoT is overly long and redundant, as it only directs the model to mimic the reasoning templates in the annotated CoT structure. Even with less optimized models like Qwen2-VL-7B, fine-tuning leads to improvements on certain benchmarks (e.g., a 5\% improvement on DynaMath and a 1.1\% improvement on LogicVista). However, for the more advanced Qwen2.5-VL-7B model, we observe significant performance degradation across all benchmarks.
(2) \textbf{\textit{Our CoT data is particularly beneficial for less optimized models.}} Specifically, for Qwen2-VL-7B, the fine-tuning results in an average improvement of 3.5\%. In contrast, applying direct SFT to the well-optimized Qwen2.5-VL-7B leads to a drop in performance.

\begin{table}[t]
\centering
\caption{The impact of RL with different reward types on all \textit{\textbf{math-type}} questions in \texttt{WeThink}.}
\resizebox{\textwidth}{!}{ 
\begin{tabular}{l | c c c c c  c |c}
\toprule[1.5pt]
Reward Type & MathVista & MathVision & MathVerse &  DynaMath & WeMath & LogicVista & Average \\ 
\midrule
Qwen2.5-VL-7B & 68.1	&25.4	&41.1	&21.8	&36.2	&47.9&  40.1 \\
Qwen2.5-VL-7B* & 69.1 & 25.9 & 41.1 & 20.8 & 35.6 & 48.5 & \underline{40.2}\\\midrule
\textit{Rule} & 65.9 & 25.1 & 42.6 & 24.0 & 39.1 & 45.2 & 40.3 (\textcolor{red}{+0.1})\\
\textit{Model} & 63.0 & 24.9 & 43.3 & 25.7 & 31.9 & 45.6 & 39.1 (\textcolor{blue}{-1.1})\\
\textit{Rule}+\textit{Model} & 66.8 & 26.2 & 45.7 & 24.2 & 37.9 & 47.4 & \textbf{41.4} (\textcolor{red}{+1.2}) \\
\bottomrule[1.5pt]
\end{tabular}
}
\label{tab:reward}
\vspace{-0.5cm}
\end{table}

\begin{table}[t]
\centering
\caption{The impact of RL with different question types (\textit{i.e.,} \textbf{\textit{math-type}} and \textbf{\textit{all-type}}) in \texttt{WeThink}, comparing performance across both mathematical and general multimodal benchmarks.}
\resizebox{\textwidth}{!}{ 
\begin{tabular}{l | c c c c c c |c}
\toprule[1.5pt]
Question Type & MathVista & MathVision & MathVerse &  DynaMath & WeMath & LogicVista &Average   \\ 
\midrule
Qwen2.5-VL-7B & 68.1	&25.4	&41.1	&21.8	&36.2	&47.9&  40.1 \\ 
Qwen2.5-VL-7B* & 69.1 & 25.9 & 41.1 & 20.8 & 35.6 & 48.5 & \underline{40.2}\\
\midrule
\textit{Math} & 66.8 & 26.2 & 45.7 & 24.2 & 37.9 & 47.4 & 41.4 (\textcolor{red}{+1.2})\\
\textit{All} & 71.6 & 26.7	& 45.1	& 24.0 & 45.5 & 	51.9	& \textbf{44.1} (\textcolor{red}{+3.9}) \\
\bottomrule[1.5pt]
\end{tabular}
}
\vspace{0.2cm}

\resizebox{\textwidth}{!}{ 
\begin{tabular}{l | c c c c c c c c |c}
\toprule[1.5pt]
Question Type &MMBench V1.1&  MMStar & MMMU   & HallusionBench & AI2D & OCRBench & MMVet & RealWorldQA &Average   \\ 
\midrule
Qwen2.5-VL-7B  & 82.2 &64.1	& 58.0   & 51.9 & 84.3	& 88.8& 69.7 & 68.4 & 70.9\\ 
Qwen2.5-VL-7B* & 82.1 & 64.5 & 57.3  & 52.9 & 84.7 & 88.2 & 67.1 & 68.2 & \underline{70.6}\\
\midrule
\textit{Math} & 79.5 & 63.3 & 57.2   & 56.0 & 83.5 & 88.3 & 72.2 &  67.8 & 71.0 (\textcolor{red}{+0.4})\\
\textit{All} & 82.0	& 64.3	& 59.3 &	55.8	& 84.3	& 88.9	& 71.7 &	67.7	& \textbf{71.8} (\textcolor{red}{+1.2}) \\
\bottomrule[1.5pt]
\end{tabular}
}
\vspace{0.2cm}
\label{tab:question_type}
\vspace{-0.8cm}
\end{table}

\begin{wrapfigure}{r}{0.4\textwidth}  
    \centering
    \includegraphics[width=0.4\textwidth]{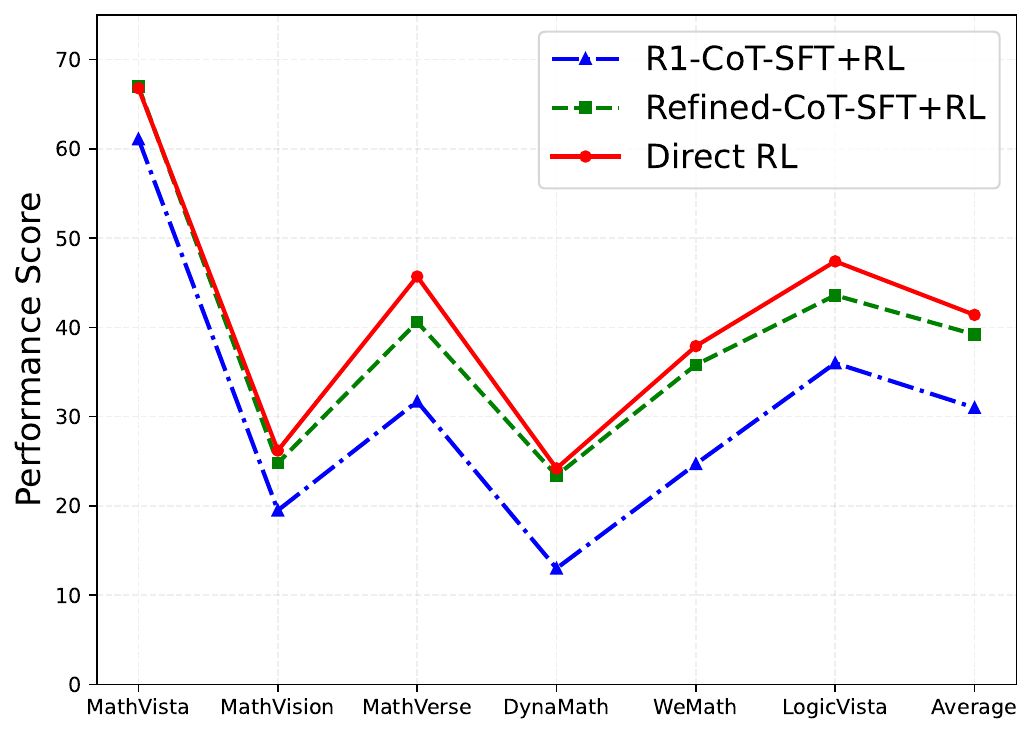}  
    \vspace{-0.4cm}
    \caption{The impact of cold start SFT for RL on all \textit{\textbf{math-type}} questions in \texttt{WeThink}.}
    \label{fig:cold_start_impact} 
    \vspace{-0.5cm}
\end{wrapfigure}

\subsection{Exploring Visual-Language Reasoning with Reinforcement Learning on \texttt{WeThink}}
\label{sec:rl_exp}
Previous experiments relying solely on CoT supervision reveal its insufficiency in exploring visual-language resoning, particularly for powerful base models. Reinforcement Learning (RL), as a more advanced training paradigm compared to CoT, has shown its effectiveness in the text-only domain, particularly with the open-source DeepSeek-R1 model. This motivates us to investigate how to harness \texttt{WeThink} for effective RL training in the visual-language context, as detailed below:

\textbf{Methods \& Implementation Details.} We adapt the DeepSeek-R1-style framework by introducing a group-relative policy optimization algorithm (GRPO)~\cite{shao2024deepseekmath} and a hybrid reward system. The hybrid reward system combines accuracy and format rewards. The accuracy reward is divided into rule-based rewards for multiple-choice and fill-in-the-blank questions in \textbf{\texttt{WeThink}}, and model-based rewards for descriptive answers, assessed using the DeepSeek-V3 judge model~\cite{liu2024deepseek}. The format reward checks for proper thinking and answer formatting in the response. The final reward combines accuracy and format rewards, with coefficients adjusting their relative importance. \textit{{\textbf{Please refer to Appendix~\ref{sec:rl} for the method details of GRPO algorithm and reward functions.}}} Our implementation is built on EasyR1~\cite{zheng2025easyr1}, which is based on veRL~\cite{sheng2024hybridflow}. We perform 5 samples per query with a temperature setting of 1.0. We select Qwen2.5-VL-7B-Instruct as our base model and perform full-parameter RL fine-tuning, with rollout and training batch sizes set to 512 and 128, respectively. For ablation studies, we use 8 NVIDIA H20s for all experiments, while 32 NVIDIA H20s is used for full-scale training on \textbf{\texttt{WeThink}}. The DeepSeek-V3 judge model is deployed on 16 NVIDIA H20s and participates in the RL training through API calls for reward computation. The system prompt is shown in Appendix Tab.~\ref{system}.

\begin{table}[t]
\centering
\caption{Comparison of different VLMs on mathematical multimodal reasoning benchmarks. We report results from other models on the OpenCompass leaderboard~\cite{duan2024vlmevalkit} and reproduce the results of our base model, Qwen2.5-VL-7B, denoted with *. $^{\dag}$ indicates the use of external images from the Internet to further enhance diversity.}
\resizebox{\textwidth}{!}{ 
\begin{tabular}{l | c c c c c  c |c}
\toprule[1.5pt]
Model & MathVista & MathVision & MathVerse &  DynaMath & WeMath & LogicVista & Average \\ 
\midrule
LLaVA-OneVision-7B & 58.6 & 18.3 & 19.3&  9.0 & 20.9 & 33.3 & 26.6 \\
InternVL2-8B &  58.3 & 20.0 & 20.4 & 9.2 & 20.2 & 33.6 & 26.9 \\
InternVL2.5-8B & 64.5 & 17.0 & 22.8 & 9.4 & 23.5 & 36.0 & 28.9 \\
Qwen2-VL-7B &61.6 &	19.2&	25.4	&11.0	&22.3&	33.3 & 28.8 \\
Qwen2.5-VL-7B & 68.1	&25.4	&41.1	&21.8	&36.2	&47.9&  40.1 \\
Qwen2.5-VL-7B* & 69.1 & 25.9 & 41.1 & 20.8 & 35.6 & 48.5 & \underline{40.2}\\
\midrule
WeThink-VL-7B & 71.6 & 26.7	& 45.1	& 24.0 & 45.5 & 	51.9	& 44.1 (\textcolor{red}{+3.9})\\
WeThink-VL-7B$^{\dag}$ & 70.9	& 27.2 &	44.7 &	24.4	& 48.0	& 53.0	& \textbf{44.7} (\textcolor{red}{+4.5})\\
\bottomrule[1.5pt]
\end{tabular}
}
\vspace{-0.5cm}
\label{tab:math}
\end{table}

\begin{table}[t]
\centering
\caption{Comparison of different VLMs on general multimodal understanding benchmarks.} 
\resizebox{\textwidth}{!}{ 
\begin{tabular}{l | c c c c c  c c c |c}
\toprule[1.5pt]
Model &MMBench V1.1&  MMStar & MMMU  & HallusionBench & AI2D & OCRBench & MMVet & RealWorldQA  &Average   \\ 
\midrule
LLaVA-OneVision-7B & 76.8	& 56.7	& 46.8	&	47.5& 82.8	&69.7	&50.6 & 69.9 &	62.6\\
InternVL2-8B & 79.4	& 61.5	& 51.2		& 45.0& 	83.6& 	79.4	& 54.3 &  64.2& 64.8 \\
InternVL2.5-8B & 82.5 & 63.2 & 56.2 	& 49.0	& 84.6 & 82.1	& 62.8	& 69.5 & 68.7 \\
Qwen2-VL-7B &  81.0	& 60.7	& 53.7		& 50.4	& 83.0	& 84.3	& 61.8& 68.5 & 67.9   \\
Qwen2.5-VL-7B & 82.2 & 64.1	& 58.0 & 51.9 & 84.3	& 88.8& 69.7 & 68.4
&  70.9\\ 
Qwen2.5-VL-7B* & 82.1 & 64.5 & 57.3  & 52.9 & 84.7 & 88.2 & 67.1 & 68.2 & \underline{70.6}\\
\midrule
WeThink-VL-7B & 82.0	& 64.3	& 59.3 &	55.8	& 84.3	& 88.9	& 71.7 &	67.7	& 71.8 (\textcolor{red}{+1.2}) \\
WeThink-VL-7B$^{\dag}$ & 82.2	& 64.8	& 61.0	& 55.5 &	84.2	& 88.8	& 73.4	& 68.2	& \textbf{72.3} (\textcolor{red}{+1.7})\\
\bottomrule[1.5pt]
\end{tabular}
}
\vspace{-0.5cm}
\label{tab:general}
\end{table}

\textbf{The impact of cold start during RL.} DeepSeek-R1 indicates that cold start SFT is instrumental in facilitating RL training. In our experiments, the previous CoT SFT model serves as the model after cold start training. 
For subsequent RL training, we select all math-type questions from \texttt{WeThink} and incorporate hybrid rewards.
As shown in Fig.~\ref{fig:sft_vs_rl_comparison}, we observe that using refined CoT as cold start data leads to substantial improvements after RL training. In contrast, applying RL to the model trained with the original R1-CoT yields minimal gains. This clearly illustrates that high-quality cold start data is critical for enhancing RL training. Furthermore, Fig.~\ref{fig:cold_start_impact} shows that applying RL directly to Qwen2.5-VL yields better improvements than the cold start method. This could be due to two factors: (1) In our scenarios, Qwen2.5-VL-7B already has strong capabilities to follow our format and requirements, which aligns with previous findings~\cite{chen2025sft}, and (2) the quality of our cold start data still requires improvement. Because the cold start SFT initially results in a performance drop, RL does provide some improvements, but the gains are not highly significant.

\textbf{The impact of RL with different reward types.} As previously demonstrated, in our scenarios, without cold start SFT, directly applying RL with hybrid rewards on math-type questions in \textbf{\texttt{WeThink}} proves to be the most effective. Here, we divide these questions into rule-based reward questions (\textit{i.e.}, multiple-choice and fill-in-the-blank questions) and model-based reward questions (\textit{i.e.} descriptive questions), and train each model for the same steps. Our results in Tab.~\ref{tab:reward} show that hybrid rewards yield the best performance, with an average improvement of 1.2\% across six benchmarks.

\textbf{The impact of RL with different question types.} 
To further improve the performance of general visual-language tasks beyond mathematics, it is crucial to integrate a wider variety of data sources that span multiple domains and contexts. Thus,
we extend our training to include all question types in \textbf{\texttt{WeThink}} for full-scale RL training with hybrid rewards. As shown in Tab.~\ref{tab:question_type}, we compare models trained on math-type questions with those trained on all question types, across 14 diverse benchmarks covering both mathematical and diverse general tasks. The results show that, firstly, the all-type question trained model significantly improves performance on mathematical reasoning benchmarks, with an average increase from 41.4\% to 44.1\%. Secondly, it substantially enhances performance  across general tasks.

\textbf{The impact of RL with increasing data diversity.}
To demonstrate the scalability of our data generation pipeline, we collect approximately 20K in-the-wild images from the Internet. Following the same data construction process, we generate new QAs for these images and incorporate them into the training. The results presented in Tab.~\ref{tab:math} and~\ref{tab:general} show that performance continues to improve across both mathematical reasoning and general benchmarks. However, the addition of new data leads to a slight decline in some benchmarks, such as MathVista and MathVerse. This suggests that RL training process is sensitive to changes in the data distribution for specific benchmarks. Nonetheless, the average improvements show our pipeline's scalability in incorporating more diverse data to further enhance model performance.

\section{Conclusion}
This work advances multimodal reasoning by tackling the challenge of general-purpose visual-language reasoning through reinforcement learning (RL). By introducing a novel \textit{Scalable Multimodal QA Synthesis} pipeline, we can generate context-aware, reasoning-centric question-answer pairs directly from the given images. We also release the \textbf{\texttt{WeThink}} dataset, containing over 120K multimodal QA pairs with annotated reasoning chains, to enhance RL training across diverse domains. We conduct comprehensive exploration of RL on our dataset, incorporating a hybrid reward mechanism, optimizes training efficiency and performance across domains. The results across 14 diverse MLLM benchmarks demonstrate the effectiveness of our dataset in improving performance from mathematical reasoning to general challenges. Furthermore, the scalability of our automated data pipeline ensures continuous improvement in data diversity, paving the way for scalable RL training across domains and general multimodal reasoning enhancement.

\noindent \textbf{Limitations.}
\textbf{\textit{Firstly}}, our proposed data pipeline relies on multiple large models, requiring more GPU resources for deployment. \textbf{\textit{Secondly}}, the pipeline generates all questions solely based on the model’s interpretation of the image content, leading to inherent randomness and instability in the generated questions. As a result, some questions need be filtered out due to quality issues. \textbf{\textit{Thirdly}}, although we employ multiple powerful models for question filtering and answer verification, our dataset still contains noise and wrong cases.

\bibliography{neurips_2025}
\bibliographystyle{unsrt}

\newpage
\appendix

\section{The Details of Scalable Multimodal QA Synthesis}
\label{sec:data_supment}

We formulate the question formulation designs into a prompt protocol for DeepSeek-R1, as illustrated in Prompt.~\ref{sec:output_specifications}, which is structured into five  parts:

\textbf{Input Information:} The input consists of potentially insufficient or erroneous preliminary visual information from Qwen2.5-VL-72B. DeepSeek-R1 relies on Qwen2.5-VL-72B to continuously refine this visual information and use it as updated input for generating reasoning-based questions through a multi-turn process.

\textbf{Core Requirements}: For each generated question, it is essential that multiple abilities are triggered, ensuring the depth and complexity of reasoning. Typically, this involves at least two reasoning steps, making the question both logical and comprehensive. Furthermore, the question design must rely on thorough image analysis to maintain clarity, depth, and completeness.

\textbf{Question Designs}: To ensure high-quality output, we have established specific guidelines for the selection of abilities. Every question must include reasoning as a mandatory component. In addition, depending on the image content, the question should also incorporate at least one of the following abilities: recognition (feature extraction), knowledge (external knowledge application), OCR (text recognition), spatial awareness (geometric or positional reasoning), or math (numerical reasoning). Moreover, we can optionally condition the generation process with contextual constraints, to achieve better control questions' type and focus.

\textbf{Information Request Protocol}: In practice, when the image description provided by Qwen2.5-VL-72B is insufficient, we activate the Information Request Protocol, which allows for up to three rounds of clarification requests to ensure that the necessary visual information is complete for generating the subsequent questions.

\textbf{Output Specifications}: Finally, the output of the question generation process must adhere to strict format specifications, including clarification requests for insufficient information (e.g., \texttt{<clarify>...</clarify>}) and properly formatted valid questions (e.g., \texttt{<q>...</q>}).

\begin{promptbox}
\subsection*{I. Input Information}
\textbf{Image Caption:} Potentially insufficient/erroneous preliminary visual information

\subsection*{II. Core Requirements}
\begin{itemize}
    \item Multi-ability triggering ($\geq$2 combinations)
    \item Explicit reasoning chains ($\geq$2 logical leaps)
    \item Comprehensive image analysis requirement
\end{itemize}

\subsection*{III. Question Design}
\begin{tcolorbox}[
    colback=white,
    colframe=gray!50,
    title=Ability Synergy Constraints,
    fonttitle=\bfseries,
    before skip=10pt,
    after skip=10pt
]
\textbf{Mandatory:} \textcolor{blue}{Reasoning}  
\\
\\
\textbf{Others (Select $\geq$1):}
\begin{tabular}{@{}ll@{}}
    \textcolor{green!50!black}{A. Recognition} & Feature extraction \& classification \\
    \textcolor{orange}{B. Knowledge} & Domain knowledge application \\
    \textcolor{red}{C. OCR} & Text recognition \& processing \\
    \textcolor{purple}{D. Spatial Awareness} &  Geometric \& positional reasoning \\
    \textcolor{cyan}{E. Math} & Numerical derivation
\end{tabular}
\end{tcolorbox}

\begin{tcolorbox}[
    colback=white,
    colframe=gray!50,
    title=Optional Contextual Constraints,
    fonttitle=\bfseries,
    before skip=10pt,
    after skip=10pt
]
    Prior Question, Task Definition, Visual Cues, etc.
\end{tcolorbox}

\subsection*{IV. Information Request Protocol}
\begin{tcolorbox}[
    colback=white,
    colframe=gray!50,
    title={Information Clarification Mechanism},
    fonttitle=\bfseries,
    before skip=10pt,
    after skip=10pt
]
 Insufficient Caption $\rightarrow$ Activate Information Request Protocol
\begin{itemize}
    \item Max conversation rounds: 3
    \item Request for more information: \texttt{<clarify>...</clarify>}
\end{itemize}
\end{tcolorbox}

\subsection*{V. Output Specifications}
\begin{tcolorbox}[
    colback=white,
    colframe=gray!50,
    title=Format Requirements,
    fonttitle=\bfseries,
    before skip=10pt,
    after skip=10pt
]
\begin{itemize}
    \item Request More Details: \texttt{<clarify>...</clarify>}
    \item Valid Question Formulation: \texttt{<q>...</q>}

\end{itemize}
\end{tcolorbox}
\label{sec:output_specifications}
\end{promptbox}

\section{Method Details}
\subsection{Problem Definition}
\label{sec:problem}
Given a multi-modal input consisting of a question \( q \) and an image \( I \), our goal is to generate the correct answer \( a \) by reasoning over both the textual and visual inputs. This reasoning process is mathematically modeled as a sequential conditional probability:
\[
P(a \mid q, I) = \prod_{t=1}^{T} P(a_t \mid q, I, a_{<t}),
\]
where \( a_t \) is the \( t \)-th token of the model's output, representing a reasoning step, and \( a_{<t} \) is the sequence of previously generated tokens. The model is expected to produce a logically consistent reasoning chain that integrates both the question and image, using elements such as mathematical formulas, contextual clues, and visual features. These reasoning steps should progressively lead to the final, accurate answer, bridging the textual and visual inputs in a structured manner.

\subsection{Supervised Fine-Tuning with Chain-of-Thought Prompting}
\label{sec:SFT}
One of the most straightforward methods to enable the model to generate explicit reasoning steps is to use chain-of-thought (CoT) prompting during supervised fine-tuning. This approach requires CoT-annotated data:
\[
D_{\text{SFT}} = \{(q_i, I_i, r_i, a_i)\},
\]
which consists of question-image pairs \( (q_i, I_i) \), along with the corresponding intermediate reasoning steps \( r_i \) and the final answer \( a_i \). The model is then trained to generate the reasoning steps \( r \) in sequence, which ultimately leads to the correct final answer. We optimize the model directly using maximum likelihood estimation for generating reasoning steps:

\[
\mathcal{L}_{\text{SFT}} = - \sum_{(q, I, r, a) \in D_{\text{SFT}}} \sum_{t=1}^{T} \log P(r_t \mid q, I, r_{<t}),
\]

where \( r_t \) represents the \( t \)-th reasoning step and \( r_{<t} \) is the sequence of previous reasoning steps. This process ensures that the model learns to generate a coherent and accurate chain of reasoning steps, which together lead to the correct final answer \( a \).

\subsection{Reinforcement Fine–tuning with Hybrid Reward}
\label{sec:rl}
While supervised fine–tuning with chain-of-thought prompting provides explicit step-by-step supervision, reinforcement learning offers a complementary paradigm for optimizing reasoning generation through reward signals. Inspired by the success of DeepSeek-R1 in text-based reasoning tasks, we adapt this framework for visual-language models by introducing a group-relative policy optimization strategy~\cite{shao2024deepseekmath} and a hybrid reward system.

\textbf{Group-Relative Policy Optimization} eliminates value function dependency through reward normalization within response groups. For each question-image pair \( (q, I) \), we sample \( G \) reasoning paths \( \{o_1, \dots, o_G\} \) from the current policy \( \pi_{\theta} \). The advantage function is computed as:
\[
\hat{A}_{i,t} = \frac{R_i - \mu(\{R_j\}_{j=1}^G)}{\sigma(\{R_j\}_{j=1}^G)}
\]
where \( \mu \) and \( \sigma \) denote the group mean and standard deviation of final rewards. The objective function combines clipped policy updates with KL regularization against the reference policy \( \pi_{\text{ref}} \):
\[
J_{\text{GRPO}}(\theta) = \mathbb{E} \left[ \frac{1}{G} \sum_{i=1}^{|o_i|} \sum_{t=1}^{|o_i|} \min \left( r_{i,t}(\theta) \hat{A}_{i,t}, \text{clip} \left( r_{i,t}(\theta), 1 - \epsilon, 1 + \epsilon \right) \hat{A}_{i,t} \right) - \beta D_{\text{KL}}(\pi_{\theta} \parallel \pi_{\text{ref}}) \right]
\]
where \( r_{i,t}(\theta) = \frac{\pi_{\theta}(o_{i,t} \mid q, I, o_{i,<t})}{\pi_{\text{old}}(o_{i,t} \mid q, I, o_{i,<t})} \) is the importance sampling ratio. This approach stabilizes training while encouraging exploration of high-reward reasoning paths.

\textbf{Hybrid Reward System} integrates accuracy and format rewards, similar to DeepSeek-R1, with accuracy reward further divided into rule-based reward and model-based reward to handle different types of answers. For example, in our \texttt{WeThink} dataset, rule-based reward is employed for multiple-choice and fill-in-the-blank questions, while model-based reward is used for descriptive questions that require longer descriptive answers.

\begin{itemize}
    \item \textbf{Rule-Based Reward} (\( R_{\text{rule}} \)): For multiple-choice and fill-in-the-blank questions, we apply exact string matching between the predicted answer and the ground truth. 
    This is done with text normalization for case and punctuation insensitivity:
    \[
    R_{\text{rule}} = \mathbb{I}(\text{normalize}(a_{\text{pred}}) = \text{normalize}(a_{\text{true}}))
    \]
    where \( \mathbb{I} \) is an indicator function, returning 1 for a true condition (exact match) and 0 for false (no match).
    
    \item \textbf{Model-Based Reward} (\( R_{\text{model}} \)): For descriptive questions, we use the DeepSeek-V3~\cite{liu2024deepseek} judge model to assess answer correctness, assigning rewards based on the clarity and correctness of the response:
    \[
    R_{\text{model}} = 
    \begin{cases}
    1 & \text{Definitely correct} \\
    0.5 & \text{Ambiguous/Partially correct} \\
    0 & \text{Definitely incorrect}
    \end{cases}
    \]
    \item \textbf{Format Reward} (\( R_{\text{format}} \)): To ensure the reasoning process is structured correctly, the format reward checks whether the response includes valid thinking and answer blocks, such as <think></think> and <answer></answer>:
    \[
    R_{\text{format}} = \mathbb{I}(\text{Valid thinking and answer blocks})
    \]
    where \( \mathbb{I} \) returns 1 for valid (correct) formatting and 0 for invalid (incorrect) formatting.
    
\end{itemize}
The final reward can be computed as:
\[
R = \alpha_{\text{accuracy}} \cdot R_{\text{accuracy}} + \alpha_{\text{format}} \cdot R_{\text{format}}
\]
where \( R_{\text{accuracy}} \) is either \( R_{\text{rule}} \) or \( R_{\text{model}} \), depending on the type of question.
The \( \alpha \) coefficients control the relative importance of the accuracy and format components. We empirically set $\alpha_{\text{accuracy}}$ to 0.7 and $\alpha_{\text{format}}$ to 0.3.

\section{Experimental Implementation Details}
\label{sec:implement_sup}

\subsection{System Prompt} To structure the reasoning process during training, we use the following system prompt for both supervised fine-tuning and reinforcement fine-tuning settings as follow:
\begin{tcolorbox}[colback=white, colframe=gray!50, title=System Prompt, fonttitle=\bfseries]
\texttt{"You FIRST think about the reasoning process as an internal monologue and then provide the final answer.
 The reasoning process MUST BE enclosed within <think> </think> tags. The final answer MUST BE enclosed within <answer> </answer> tags."}
 \label{system}
\end{tcolorbox}

\subsection{Benchmarks \& Evaluation.} 
\label{sec:eval}
As illustrated in Sec.~\ref{sec:benchmarks}, we conduct comprehensive evaluation across 14 MLLM benchmarks, covering six mathematical reasoning benchmarks and eight general multimodal understanding benchmarks. Below are the details:

\textbf{Mathematical reasoning} employs the following benchmarks: 
$\diamond$  {MathVista\_MINI}, which is the Test Mini split of the MathVista dataset

$\diamond$ {MathVision}, which uses the full test set of MathVision

$\diamond$ {MathVerse\_MINI\_Vision\_Only}, the Test Mini split of MathVerse, using the "Vision Only" mode

$\diamond$ {DynaMath}, which uses the full test set of DynaMath

$\diamond$ {WeMath}, the Test Mini split of WeMath, where we report "Score (Strict)" as the main metric

$\diamond$ {LogicVista}, which is the full test set of LogicVista

\textbf{General multimodal understanding} employs MMBench v1.1 (Test\_CN / Test\_EN), MMStar, MMMU (Val), HallusionBench, AI2D (Test), OCRBench, MMVet and RealWorldQA.

\textbf{Evaluation Details.} To ensure fairness and reproducibility, we conduct evaluation of our models using VLMEvalKit~\cite{duan2024vlmevalkit}, an open-source toolkit designed for MLLM evaluation. All evaluations are conducted using 8 NVIDIA A800s. We follow the Qwen2.5-VL series settings, where the minimum number of pixels is set to 1280*28*28, the maximum number of pixels is set to 16384*28*28, and the \texttt{use\_custom\_prompt} option is disabled. The \texttt{max\_new\_tokens} parameter is set to 2048 by default, in alignment with Qwen2.5-VL series. The system prompt is provided in Tab.~\ref{system}. We empirically found that using the default system prompt of the Qwen2.5-VL series yields better performance for OCRBench and AI2D benchmarks.

\section{Case Study: From Math to General Challenges}
Below, we conduct case study to investigate how reasoning contributes to performance improvements in diverse scenarios，
including \textit{Mathematics}, \textit{Medicine}, \textit{Humanities}, \textit{Science}, \textit{Art}, \textit{Engineering} and \textit{Daily Scenes}. These examples show that WeThink-VL-7B exhibits superior logical and analytical capabilities, leading to correct answer.

\begin{tcolorbox}[title = {Mathematics}, breakable, fontupper=\small, fonttitle=\small]

\vspace{\medskipamount}
{\includegraphics[width=0.5\linewidth]{figures/case_study/1.jpg}}
\vspace{\medskipamount}

Question: The casino game, roulette, allows the gambler to bet on the probability of a ball, which spins in the roulette wheel, landing on a particular color, number, or range of numbers. The table used to place bets contains of 38 numbers, and each number is assigned to a color and a range.  Compute the probability of winning the following type of bets: Betting on three numbers in a line, as in 1-2-3

Options: 

A. 1/38

B. 2/38

C. 3/38

D. 6/38

GT Answer: C

\noindent\rule{\linewidth}{0.5pt}
\textbf{Qwen2.5-VL-7B}

To determine the probability of winning a bet on three numbers in a line (such as 1-2-3) in roulette, we need to understand the total number of possible outcomes and the number of favorable outcomes.

1. Total number of possible outcomes: There are 38 numbers on the roulette wheel.

2. Favorable outcomes: There is only 1 way to win by betting on the exact three numbers 1-2-3.

The probability \( P \) of winning is given by the ratio of the number of favorable outcomes to the total number of possible outcomes:

\[ P = \frac{\text{Number of favorable outcomes}}{\text{Total number of possible outcomes}} = \frac{1}{38} \]

Therefore, the correct answer is:

\textcolor{blue}{A. 1/38}

\noindent\rule{\linewidth}{0.5pt}
\textbf{WeThink-VL-7B: }

<think>
To compute the probability of winning by betting on three numbers in a line, we need to understand that there are 38 possible numbers on the roulette wheel. When you bet on three specific numbers, you are essentially looking for the probability that the ball lands on one of those three numbers.

The probability of the ball landing on any specific number is 1/38. Since you are betting on three specific numbers, the probability of winning is the sum of the probabilities of the ball landing on each of those three numbers. However, since these are mutually exclusive events (the ball can't land on more than one number at a time), we simply add the probabilities:

\[ \text{Probability} = \frac{1}{38} + \frac{1}{38} + \frac{1}{38} = \frac{3}{38} \]

So, the probability of winning by betting on three numbers in a line is 3/38.
</think>

<answer>
\textcolor{red}{C. 3/38}
</answer>

\end{tcolorbox}

\begin{tcolorbox}[title = {Medicine}, breakable, fontupper=\small, fonttitle=\small]

\vspace{\medskipamount}
{\includegraphics[width=\linewidth]{figures/case_study/2.jpg}}
\vspace{\medskipamount}

Question: What is this rhythm?

Options:

A. Normal Sinus Rhythm

B. Sinus Bradycardia

C. Accelerated Junctional Rhythm

D. NSR with 1st Degree AV Block

E. Idioventricular Rhythm

F. Paced with Failure to Capture

GT Answer: C

\noindent\rule{\linewidth}{0.5pt}

\textbf{Qwen2.5-VL-7B}

The rhythm in the provided ECG appears to be irregular with a slow heart rate. The P waves are not clearly visible, and the QRS complexes are wide and irregular. This suggests a rhythm that is not sinus rhythm. Given the options provided, the most likely diagnosis is:

\textcolor{blue}{E. Idioventricular Rhythm}

Idioventricular rhythm is characterized by wide, irregular QRS complexes and a slow heart rate, which matches the appearance of the rhythm in the ECG.

\noindent\rule{\linewidth}{0.5pt}

\textbf{WeThink-VL-7B: }

<think>The image provided is an ECG (Electrocardiogram) which is used to monitor the electrical activity of the heart. The rhythm in the ECG shows regular, narrow QRS complexes with no P waves preceding them, which is characteristic of junctional rhythms. The QRS complexes are regular and the heart rate appears to be relatively slow, which is not typical for normal sinus rhythm or bradycardia. The absence of P waves and the regularity of the QRS complexes suggest a junctional rhythm. The regularity and the narrow QRS complexes also rule out idioventricular rhythm, which is typically irregular and wide QRS complexes. The paced rhythm with failure to capture is not applicable as there are no artificial pacing spikes visible. The rhythm is not consistent with first-degree AV block as the QRS complexes are narrow and regular. Therefore, the most likely rhythm is an accelerated junctional rhythm.
</think>

<answer>
\textcolor{red}{C. Accelerated Junctional Rhythm}
</answer>

\end{tcolorbox}

\begin{tcolorbox}[title = {Humanities}, breakable, fontupper=\small, fonttitle=\small]

\vspace{\medskipamount}
{\includegraphics[width=0.5\linewidth]{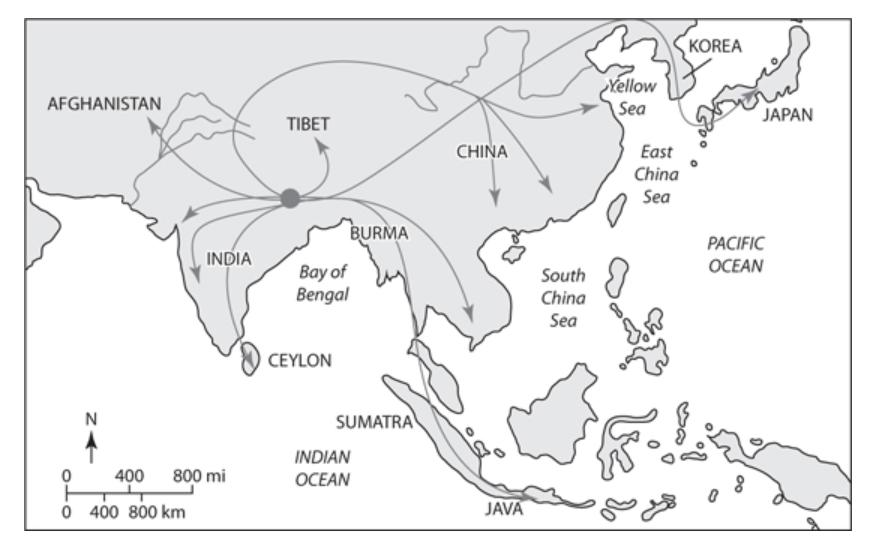}}
\vspace{\medskipamount}

Question: A historian would use the above map to illustrate which of the following?

Options: 

A. Major trade routes in the classical era

B. The spread of communism from 1900 to today

C. One aspect of cultural diffusion

D. The monsoonal wind patterns of Asia

GT Answer: C

\noindent\rule{\linewidth}{0.5pt}
\textbf{Qwen2.5-VL-7B}

The correct answer is:

\textcolor{blue}{A. Major trade routes in the classical era}

The map shows various routes connecting different regions, which is indicative of trade and cultural exchange patterns. The presence of arrows and the geographical context suggest the movement of goods and ideas, which is characteristic of major trade routes in the classical era.

\noindent\rule{\linewidth}{0.5pt}
\textbf{WeThink-VL-7B: }

<think>
The map shows various routes connecting different parts of Asia, including China, India, and the islands of Southeast Asia. These routes are likely to represent trade or cultural exchanges rather than wind patterns or the spread of communism. The presence of arrows and the interconnected nature of the lines suggest movement and exchange, which is characteristic of trade routes or cultural diffusion. The map does not specifically indicate monsoonal wind patterns, as it lacks the necessary meteorological data and is more focused on geographical and cultural connections.
</think>

<answer>
\textcolor{red}{C. One aspect of cultural diffusion}
</answer>
\end{tcolorbox}

\begin{tcolorbox}[title = {Science}, breakable, fontupper=\small, fonttitle=\small]

\vspace{\medskipamount}
{\includegraphics[width=0.3\linewidth]{figures/case_study/4.jpg}}
\vspace{\medskipamount}

Question: In the Section of left leg, identify the 170 structure. 

Options: 

A. Tibialis anterior

B. Tibialis posterior

C. Flexor hallucis longus

D. Peroneus longus

GT Answer: D

\noindent\rule{\linewidth}{0.5pt}
\textbf{Qwen2.5-VL-7B}

The structure labeled as 170 in the image appears to be the tibialis posterior muscle, which is located in the posterior compartment of the leg. Therefore, the correct answer is:

\textcolor{blue}{B. Tibialis posterior}

\noindent\rule{\linewidth}{0.5pt}
\textbf{WeThink-VL-7B: }

<think>
The image provided is a cross-sectional view of a leg, likely representing the anatomy of the lower leg. The structure labeled as 170 appears to be located on the outer side of the section, which corresponds to the lateral side of the leg. In the anatomy of the leg, the tibialis posterior is typically located on the medial side, the tibialis anterior on the anterior side, the flexor hallucis longus is deep within the leg, and the peroneus longus is on the lateral side, which matches the location of the structure labeled 170.
</think>

<answer>
\textcolor{red}{The correct answer is D. Peroneus longus.}
</answer>

\end{tcolorbox}

\begin{tcolorbox}[title = {Art}, breakable, fontupper=\small, fonttitle=\small]

\vspace{\medskipamount}
{\includegraphics[width=0.3\linewidth]{figures/case_study/5.jpg}}
\vspace{\medskipamount}

Question: The painting shown was created by:

Options: 

A. Wifredo Lam

B. Frida Kahlo

C. Diego Rivera

D. Jacob Lawrence

GT Answer: D

\noindent\rule{\linewidth}{0.5pt}
\textbf{Qwen2.5-VL-7B:}

 The painting shown is "The Four Stages of a Man" by Wifredo Lam. Therefore, the correct answer is:

\textcolor{blue}{A. Wifredo Lam}

\noindent\rule{\linewidth}{0.5pt}
\textbf{WeThink-VL-7B: }

<think>
The image depicts a painting with a series of figures engaged in various activities, such as reading, writing, and working. The style of the painting, with its flat, almost cut-out appearance of the figures and the use of a single, prominent line that seems to connect the different scenes, is characteristic of the work of Jacob Lawrence. Jacob Lawrence is known for his series of paintings that often depict historical events or social issues, and his style often includes a similar use of bold lines and a focus on the human figure. The other artists listed have distinct styles: Wifredo Lam is known for his surrealist works, Frida Kahlo for her self-portraits and symbolic works, and Diego Rivera for his large-scale murals and political themes. The style of this painting does not align with the known styles of the other artists.
</think>

<answer>
\textcolor{red}{D. Jacob Lawrence}
</answer>

\end{tcolorbox}

\begin{tcolorbox}[title = {Engineering}, breakable, fontupper=\small, fonttitle=\small]

\vspace{\medskipamount}
{\includegraphics[width=0.5\linewidth]{figures/case_study/6.jpg}}
\vspace{\medskipamount}

Question: The adjacency list of a directed graph is given as follows. Starting from the vertex V1 and traversing according to the depth-first search method, the obtained vertex sequence is:

Options: 

A. V1,V5,V4,V7,V6,V2,V3

B. V1,V2,V3,V4,V7,V6,V5

C. V1,V5,V4,V7,V6,V3,V2

D. V1,V5,V6,V4,V7,V2,V3

GT Answer: C

\noindent\rule{\linewidth}{0.5pt}
\textbf{Qwen2.5-VL-7B:}

To determine the correct vertex sequence obtained by traversing the graph using a depth-first search (DFS) starting from vertex V1, we need to follow the adjacency list and the DFS rules:

1. Start at V1.

2. Visit V1 and then follow the edges to its neighbors.

3. Continue this process, visiting each vertex and its neighbors, ensuring that each vertex is visited only once.

Let's follow the adjacency list step-by-step:

1. Start at V1.

2. V1 points to V5, so visit V5.

3. V5 points to V4, so visit V4.

4. V4 points to V7, so visit V7.

5. V7 points to V6, so visit V6.

6. V6 has no further neighbors, so backtrack to V5.

7. Since V5 has already been visited, backtrack to V4.

8. Since V4 has already been visited, backtrack to V1.

9. Since V1 has already been visited, backtrack to V2.

10. V2 has no further neighbors, so backtrack to V3.

11. V3 has no further neighbors, so backtrack to V1.

The sequence of vertices visited is: V1, V5, V4, V7, V6, V2, V3.

Thus, the correct answer is:
\textcolor{blue}{A. V1,V5,V4,V7,V6,V2,V3}

\noindent\rule{\linewidth}{0.5pt}
\textbf{WeThink-VL-7B: }

 <think>
To determine the correct vertex sequence obtained by traversing the graph from vertex V1 using a depth-first search (DFS) method, we need to follow the adjacency list and the rules of DFS. DFS typically involves visiting a vertex, then exploring as far as possible along each branch before backtracking.

Starting from V1:

1. V1 -> V5 (first child)

2. V5 -> V4 (first child)

3. V4 -> V7 (first child)

4. V7 -> V6 (first child)

5. V6 -> (no further children, backtrack)

6. V4 -> (already visited, backtrack)

7. V5 -> V3 (next child)

8. V3 -> V2 (first child)

9. V2 -> (no further children, backtrack)

10. V3 -> (already visited, backtrack)

11. V5 -> (already visited, backtrack)

12. V1 -> (already visited, backtrack)

The sequence obtained is: V1, V5, V4, V7, V6, V3, V2.

Let's match this with the given options:

A. V1,V5,V4,V7,V6,V2,V3

B. V1,V2,V3,V4,V7,V6,V5

C. V1,V5,V4,V7,V6,V3,V2

D. V1,V5,V6,V4,V7,V2,V3

The correct sequence is: V1, V5, V4, V7, V6, V3, V2, which matches option C.
</think>

<answer>
\textcolor{red}{C. V1,V5,V4,V7,V6,V3,V2}
</answer>

\end{tcolorbox}

\begin{tcolorbox}[title = {Daily Scenes}, breakable, fontupper=\small, fonttitle=\small]

\vspace{\medskipamount}
{\includegraphics[width=0.4\linewidth]{figures/case_study/8.png}}
\vspace{\medskipamount}

Question: What fruit is the right of plums?

GT Answer: orange

\noindent\rule{\linewidth}{0.5pt}
\textbf{Qwen2.5-VL-7B}

\textcolor{blue}{The fruit to the right of the plums in the image is a kiwi.}

\noindent\rule{\linewidth}{0.5pt}
\textbf{WeThink-VL-7B: }

<think>
The image shows a variety of fruits and vegetables. The plums are located in the middle of the image, slightly to the left. To the right of the plums, there are several oranges. The oranges are clearly visible and are positioned next to the plums.
</think>

<answer>
\textcolor{red}{The fruit to the right of the plums is an orange.}
</answer>

\vspace{\medskipamount}
{\includegraphics[width=0.6\linewidth]{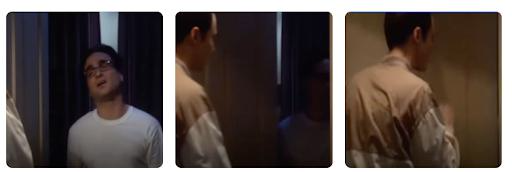}}
\vspace{\medskipamount}

Question: According to the positive sequence of the images, does the man close the door?

GT Answer: Yes

\noindent\rule{\linewidth}{0.5pt}
\textbf{Qwen2.5-VL-7B}

\textcolor{blue}{No, the man does not close the door in the sequence of images provided.} The images show the man standing in front of a door, but there is no indication that he closes the door. The sequence appears to show the man facing the door, possibly preparing to open it, but the door is not closed in the images.

\noindent\rule{\linewidth}{0.5pt}
\textbf{WeThink-VL-7B: }

<think>
The sequence of images shows a man in a white shirt. In the first image, he is facing someone or something off-camera. In the second image, he is seen from the side, and it appears he is moving towards a door. In the third image, he is seen from behind, and it looks like he is interacting with the door, possibly closing it. The progression of the images suggests a movement towards the door and a possible action of closing it.
</think>

<answer>
\textcolor{red}{Yes, according to the positive sequence of the images, the man appears to close the door.}
</answer>

\end{tcolorbox}

\end{document}